\documentclass[11pt,letterpaper]{article}
\usepackage{emnlp2016}
\usepackage{times}
\usepackage{latexsym}
\usepackage[pass]{geometry}
\usepackage{amsfonts}
\usepackage{amsthm,amsmath}
\usepackage{amssymb}
\usepackage{fixltx2e}
\usepackage{graphicx}
\usepackage{url}

\emnlpfinalcopy

\def\emnlppaperid{***}

\title{Bridging the Gap: Incorporating a Semantic Similarity Measure for Effectively Mapping PubMed Queries to Documents}

\author{Sun Kim, Nicolas Fiorini, W. John Wilbur and Zhiyong Lu \\
  National Center for Biotechnology Information \\
  National Library of Medicine, National Institutes of Health \\
  Bethesda, MD 20894, USA \\
  {\tt \{sun.kim,nicolas.fiorini,john.wilbur,zhiyong.lu\}@nih.gov} \\}

\date{}

\begin{document}

\maketitle

\begin{abstract}
The main approach of traditional information retrieval (IR) is to
examine how many words from a query appear in a document. A
drawback of this approach, however, is that it may fail to detect
relevant documents where no or only few words from a query are
found. The semantic analysis methods such as LSA (latent semantic
analysis) and LDA (latent Dirichlet allocation) have been proposed
to address the issue, but their performance is not superior
compared to common IR approaches. Here we present a query-document
similarity measure motivated by the Word Mover's Distance. Unlike
other similarity measures, the proposed method relies on neural
word embeddings to compute the distance between words. This
process helps identify related words when no direct matches are
found between a query and a document. Our method is efficient and
straightforward to implement. The experimental results on TREC
Genomics data show that our approach outperforms the BM25 ranking
function by an average of 12\% in mean average precision.
Furthermore, for a real-world dataset collected from the
PubMed\textsuperscript{\textregistered} search logs, we combine
the semantic measure with BM25 using a learning to rank method,
which leads to improved ranking scores by up to 25\%. This
experiment demonstrates that the proposed approach and BM25 nicely
complement each other and together produce superior performance.
\end{abstract}

\section{Introduction}

In information retrieval (IR), queries and documents are typically
represented by term vectors where each term is a content word and
weighted by \textit{tf-idf}, i.e. the product of the term frequency
and the inverse document frequency, or other weighting schemes
\cite{Salton1988}. The similarity of a query and a document is then
determined as a dot product or cosine similarity. Although this
works reasonably, the traditional IR scheme often fails to find
relevant documents when synonymous or polysemous words are used in a
dataset, e.g. a document including only ``neoplasm" cannot be found
when the word ``cancer" is used in a query. One solution of this
problem is to use query expansion
\cite{Lu2009,Carpineto2012,Diaz2016,Roy2016} or dictionaries, but
these alternatives still depend on the same philosophy, i.e. queries
and documents should share exactly the same words.

While the term vector model computes similarities in a sparse and
high-dimensional space, the semantic analysis methods such as latent
semantic analysis (LSA) \cite{Deerwester1990,Hofmann1999} and latent
Dirichlet allocation (LDA) \cite{Blei2003} learn dense vector
representations in a low-dimensional space. These methods choose a
vector embedding for each term and estimate a similarity between
terms by taking an inner product of their corresponding embeddings
\cite{Sordoni2014}. Since the similarity is calculated in a latent
(semantic) space based on context, the semantic analysis approaches
do not require having common words between a query and documents.
However, it has been shown that LSA and LDA methods do not produce
superior results in various IR tasks
\cite{Maas2011,Baroni2014,Pennington2014} and the classic ranking
method, BM25 \cite{Robertson2009}, usually outperforms those methods
in document ranking \cite{Atreya2011,Nalisnick2016}.

Neural word embedding \cite{Bengio2003,Mikolov2013} is similar to
the semantic analysis methods described above. It learns
low-dimensional word vectors from text, but while LSA and LDA
utilize co-occurrences of words, neural word embedding learns word
vectors to predict context words \cite{Baroni2014}. Moreover,
training of semantic vectors is derived from neural networks. Both
co-occurrence and neural word embedding approaches have been used
for lexical semantic tasks such as semantic relatedness (e.g. king
and queen), synonym detection (e.g. cancer and carcinoma) and
concept categorization (e.g. banana and pineapple belong to fruits)
\cite{Baroni2014,Schnabel2015}. But, Baroni et al.
\shortcite{Baroni2014} showed that neural word embedding approaches
generally performed better on such tasks with less effort required
for parameter optimization. The neural word embedding models have
also gained popularity in recent years due to their high performance
in NLP tasks \cite{Levy2014}.

Here we present a query-document similarity measure using a neural
word embedding approach. This work is particularly motivated by the
Word Mover's Distance \cite{Kusner2015}. Unlike the common
similarity measure taking query/document centroids of word
embeddings, the proposed method evaluates a distance between
individual words from a query and a document. Our first experiment
was performed on the TREC 2006 and 2007 Genomics benchmark sets
\cite{Hersh2006,Hersh2007}, and the experimental results showed that
our approach was better than BM25 ranking. This was solely based on
matching queries and documents by the semantic measure and no other
feature was used for ranking documents.

In general, conventional ranking models (e.g. BM25) rely on a
manually designed ranking function and require heuristic
optimization for parameters \cite{Liu2009,Chapelle2011}. In the age
of information explosion, this one-size-fits-all solution is no
longer adequate. For instance, it is well known that links to a web
page are an important source of information in web document search
\cite{Brin1998}, hence using the link information as well as the
relevance between a query and a document is crucial for better
ranking. In this regard, learning to rank \cite{Liu2009} has drawn
much attention as a scheme to learn how to combine diverse features.
Given feature vectors of documents and their relevance levels, a
learning to rank approach learns an optimal way of weighting and
combining multiple features.

We argue that the single scores (or features) produced by BM25 and
our proposed semantic measure complement each other, thus merging
these two has a synergistic effect. To confirm this, we measured the
impact on document ranking by combining BM25 and semantic scores
using the learning to rank approach, LamdaMART
\cite{Burges2008,Burges2010}. Trained on PubMed user queries and
their click-through data, we evaluated the search performance based
on the most highly ranked 20 documents. As a result, we found that
using our semantic measure further improved the performance of BM25.

Taken together, we make the following important contributions in
this work. First, to the best of our knowledge, this work represents
the first investigation of query-document similarity for information
retrieval using the recently proposed Word Mover's Distance. Second,
we modify the original Word Mover's Distance algorithm so that it is
computationally less expensive and thus more practical and scalable
for real-world search scenarios (e.g. biomedical literature search).
Third, we measure the actual impact of neural word embeddings in
PubMed by utilizing user queries and relevance information derived
from click-through data. Finally, on TREC and PubMed datasets, our
proposed method achieves stronger performance than BM25.

\section{Methods}

A common approach to computing similarity between texts (e.g.
phrases, sentences or documents) is to take a centroid of word
embeddings, and evaluate an inner product or cosine similarity
between centroids\footnote{The implementation of \textit{word2vec}
also uses centroids of word vectors for calculating similarities
(\url{https://code.google.com/archive/p/word2vec}).}
\cite{Nalisnick2016,Furnas1988}. This has found use in
classification and clustering because they seek an overall topic of
each document. However, taking a simple centroid is not a good
approximator for calculating a distance between a query and a
document \cite{Kusner2015}. This is mostly because queries tend to
be short and finding the actual query words in documents is feasible
and more accurate than comparing lossy centroids. Consistent with
this, our approach here is to measure the distance between
individual words, not the average distance between a query and a
document.

\subsection{Word Mover's Distance}

Our work is based on the Word Mover's Distance between text
documents \cite{Kusner2015}, which calculates the minimum cumulative
distance that words from a document need to travel to match words
from a second document. In this subsection, we outline the original
Word Mover's Distance algorithm, and our adapted model is described
in Section 2.2.

First, following Kusner et al. \shortcite{Kusner2015}, documents are
represented by normalized bag-of-words (BOW) vectors, i.e. if a word
$w_i$ appears $tf_i$ times in a document, the weight is
\begin{eqnarray}
d_i = \frac{tf_i}{\sum_{i'=1}^n tf_{i'}},
\end{eqnarray}
\noindent where $n$ is number of words in the document. The higher
the weight, the more important the word. They assume a word
embedding so that each word $w_i$ has an associated vector
$\mathbf{x}_i$. The dissimilarity $c$ between $w_i$ and $w_j$ is
then calculated by
\begin{eqnarray}
c(i,j)=\|\mathbf{x}_i-\mathbf{x}_j\|_2.
\end{eqnarray}
\noindent The Word Mover's Distance makes use of word importance and
the relatedness of words as we now describe.

Let $\mathbf{D}$ and $\mathbf{D'}$ be BOW representations of two
documents $D$ and $D'$. Let $\mathbf{T} \in \mathcal{R}^{n \times
n}$ be a flow matrix, where $\mathbf{T}_{ij} \geq 0$ denotes how
much it costs to travel from $w_i$ in $D$ to $w_j$ in $D'$, and $n$
is the number of unique words appearing in $D$ and/or $D'$. To
entirely transform $\mathbf{D}$ to $\mathbf{D'}$, we ensure that the
entire outgoing flow from $w_i$ equals $d_i$ and the incoming flow
to $w_j$ equals $d'_j$. The Word Mover's Distance between $D$ and
$D'$ is then defined as the minimum cumulative cost required to move
all words from $D$ to $D'$ or vice versa, i.e.
\begin{eqnarray}
\min_{\mathbf{T} \geq 0} && \sum_{i,j=1}^n \mathbf{T}_{ij} c(i,j) \label{eq:01} \\
\mathrm{subject~to} && \sum_{j=1}^n \mathbf{T}_{ij} = d_i, \forall i \in \{1,...,n\} \nonumber \\
&& \sum_{i=1}^n \mathbf{T}_{ij} = d'_j, \forall j \in \{1,...,n\}. \nonumber
\end{eqnarray}
The solution is attained by finding $\mathbf{T}_{ij}$ that minimizes
the expression in Eq. (\ref{eq:01}). Kusner et al.
\shortcite{Kusner2015} applied this to obtain nearest neighbors for
document classification, i.e. \textit{k}-NN classification and it
produced outstanding performance among other state-of-the-art
approaches. What we have just described is the approach given in
Kusner et al. We will modify the word weights and the measure of the
relatedness of words to better suit our application.

\subsection{Our Query-Document Similarity Measure}

While the prior work gives a hint that the Word Mover's Distance is
a reasonable choice for evaluating a similarity between documents,
it is uncertain how the same measure could be used for searching
documents to satisfy a query. First, it is expensive to compute the
Word Mover's Distance. The time complexity of solving the distance
problem is $O(n^3 \log n)$ \cite{Pele2009}. Second, the semantic
space of queries is not the same as those of documents. A query
consists of a small number of words in general, hence words in a
query tend to be more ambiguous because of the restricted context.
On the contrary, a text document is longer and more informational.
Having this in mind, we realize that ideally two distinctive
components could be employed for query-document search: 1) mapping
queries to documents using a word embedding model trained on a
document set and 2) mapping documents to queries using a word
embedding model obtained from a query set. In this work, however, we
aim to address the former, and the mapping of documents to queries
remains as future work.

For our purpose, we will change the word weight $d_i$ to incorporate
inverse document frequency ($idf$), i.e.
\begin{eqnarray}
d_i = idf(i) \frac{tf_i}{\sum_{i'=1}^n tf_{i'}},
\label{eq:03}
\end{eqnarray}
\noindent where $idf(i) = \log \frac{K-k_i+0.5}{k_i+0.5}$. $K$ is
the size of a document set and $k_i$ is the number of documents that
include the $i$th term. The rationale behind this is to weight words
in such a way that common terms are given less importance. It is the
\textit{idf} factor normally used in \textit{tf-idf} and BM25
\cite{Witten1999,Wilbur2001}. In addition, our word embedding is a
neural word embedding trained on the 25 million PubMed titles and
abstracts.

Let $\mathbf{Q}$ and $\mathbf{D}$ be BOW representations of a query
$Q$ and a document $D$. $D$ and $D'$ in Section 2.1 are now replaced
by $Q$ and $D$, respectively. Since we want to have a higher score
for documents relevant to $Q$, $c(i,j)$ is redefined as a cosine
similarity, i.e.
\begin{eqnarray}
c(i,j)=\frac{\mathbf{x}_i \cdot
\mathbf{x}_j}{\|\mathbf{x}_i\|\|\mathbf{x}_j\|}.
\end{eqnarray}
In addition, the problem we try to solve is the flow $Q \rightarrow
D$. Hence, Eq. (\ref{eq:01}) is rewritten as follows.
\begin{eqnarray}
\max_{\mathbf{T} \geq 0} && \sum_{i,j=1}^n \mathbf{T}_{ij} c(i,j) \label{eq:02} \\
\mathrm{subject~to} && \sum_{j=1}^n \mathbf{T}_{ij} = d_i, \forall i \in \{1,...,n\}, \nonumber
\end{eqnarray}
\noindent where $d_i$ represents the word $w_i$ in $Q$. $idf(i)$ in
Eq. (\ref{eq:03}) is unknown for queries, therefore we compute
$idf(i)$ based on the document collection. The optimal solution of
the expression in Eq. (\ref{eq:02}) is to map each word in $Q$ to
the most similar word in $D$ based on word embeddings. The time
complexity for getting the optimal solution is $O(mn)$, where $m$ is
the number of unique query words and $n$ is the number of unique
document words. In general, $m \ll n$ and evaluating the similarity
between a query and a document can be implemented in parallel
computation. Thus, the document ranking process can be quite
efficient.

\subsection{Learning to Rank}

In our study, we use learning to rank to merge two distinctive
features, BM25 scores and our semantic measures. This approach is
trained and evaluated on real-world PubMed user queries and their
responses based on click-through data \cite{Joachims2002a}. While it
is not common to use only two features for learning to rank, this
approach is scalable and versatile. Adding more features
subsequently should be straightforward and easy to implement. The
performance result we obtain demonstrates the semantic measure is
useful to rank documents according to users' interests.

We briefly outline learning to rank approaches
\cite{Severyn2015,Freund2003} in this subsection. For a list of
retrieved documents, i.e. for a query $Q$ and a set of candidate
documents, $D = \{ D_1, D_2, ..., D_m \}$, we are given their
relevancy judgements $y = \{ y_1, y_2, ..., y_m \}$, where $y_i$ is
a positive integer when the document $D_i$ is relevant and 0
otherwise. The goal of learning to rank is to build a model $h$ that
can rank relevant documents near or at the top of the ranked
retrieval list. To accomplish this, it is common to learn a function
$h (\textbf{w}, \psi (Q,D))$, where $\textbf{w}$ is a weight vector
applied to the feature vector $\psi (Q,D)$. A part of learning
involves learning the weight vector but the form of $h$ may also
require learning. For example, $h$ may involve learned decision
trees as in our application.

In particular, we use LambdaMART \cite{Burges2008,Burges2010} for
our experiments. LambdaMART is a pairwise learning to rank approach
and is being used for PubMed relevance search. While the simplest
approach (pointwise learning) is to train the function $h$ directly,
pairwise approaches seek to train the model to place correct pairs
higher than incorrect pairs, i.e. $ h (\textbf{w}, \psi (Q,D_i))
\geq h (\textbf{w}, \psi (Q,D_j)) + \epsilon $, where the document
$D_i$ is relevant and $D_j$ is irrelevant. $\epsilon$ indicates a
margin. LambdaMART is a boosted tree version of LambdaRank
\cite{Burges2010}. An ensemble of LambdaMART, LambdaRank and
logistic regression models won the Yahoo! learning to rank challenge
\cite{Chapelle2011}.

\section{Results and Discussion}

\begin{figure*}[!tpb]
\begin{center}
\fbox{\includegraphics[angle=0,width=0.9\textwidth]{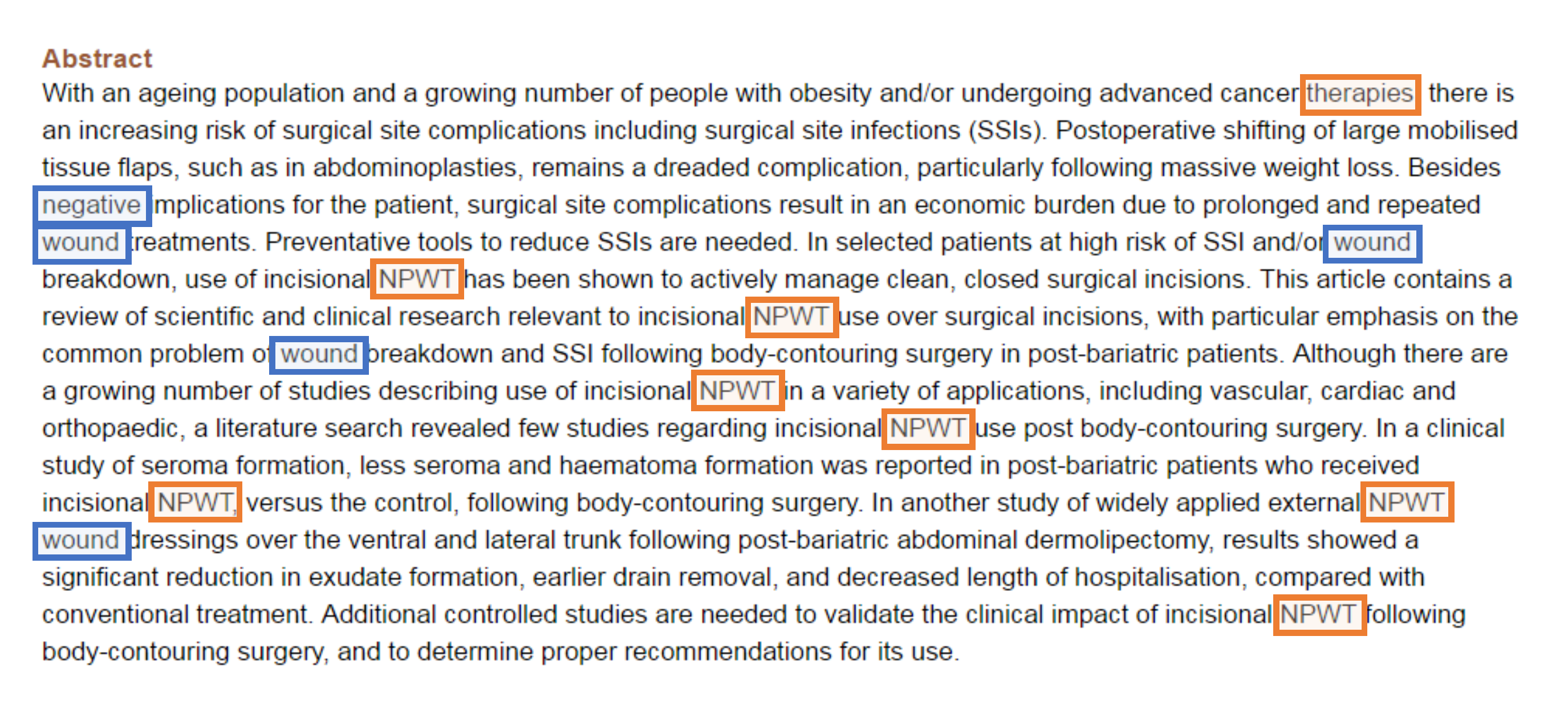}}
\end{center}
\caption{Matching the query, ``negative pressure wound therapy" with
the words in the abstract of PMID 25853645 using the exact string match vs.
the semantic string match.
The blue boxes indicate that the words appear in the query.
Additional words (orange boxes) are found using our semantic measure.}
\label{fig:01}
\end{figure*}

Our resulting formula from the Word Mover's Distance seeks to find
the closest terms for each query word. Figure \ref{fig:01} depicts
an example with and without using our semantic matching. For the
query, ``negative pressure wound therapy", a traditional way of
searching documents is to find those documents which include the
words ``negative", ``pressure", ``wound" and ``therapy". As shown in
the figure, the words, ``pressure" and ``therapy", cannot be found
by perfect string match. On the other hand, within the same context,
the semantic measure finds the closest words ``NPWT" and
``therapies" for ``pressure" and ``therapy", respectively.
Identifying abbreviations and singular/plural would help match the
same words, but this example is to give a general idea about the
semantic matching process. Also note that using dictionaries such as
synonyms and abbreviations requires an additional effort for manual
annotation.

In the following subsections, we describe the datasets and
experiments, and discuss our results.

\subsection{Datasets}

To evaluate our word embedding approach, we used two scientific
literature datasets: TREC Genomics data and PubMed. Table
\ref{tab:01} shows the number of queries and documents in each
dataset. TREC represents the benchmark sets created for the TREC
2006 and 2007 Genomics Tracks \cite{Hersh2006,Hersh2007}. The
original task is to retrieve passages relevant to topics (i.e.
queries) from full-text articles, but the same set can be utilized
for searching relevant PubMed documents. We consider a PubMed
document relevant to a TREC query if and only if the full-text of
the document contains a passage judged relevant to that query by the
TREC judges. Our setup is more challenging because we only use
PubMed abstracts, not full-text articles, to find evidence.

\begin{table}
\centering
\begin{tabular}{cccc}
\hline \bf Dataset & \bf \# Queries & \bf \# Documents \\ \hline
TREC 2006 & 26 & 162,259 \\
TREC 2007 & 36 & 162,259 \\
PubMed & 27,870 & 27,098,629\footnotemark\\
\hline
\end{tabular}
\caption{Number of queries and documents for TREC and PubMed experiments.
TREC 2006 includes 28 queries originally but two were removed because there were no relevant documents.\label{tab:01}}
\end{table}
\footnotetext{This is the number of PubMed documents as of Apr. 6,
2017. This number and the actual number of documents used for our
experiments may differ slightly.}

Machine learning approaches, especially supervised ones such as
learning to rank, are promising and popular nowadays. Nonetheless,
they usually require a large set of training examples, and such
datasets are particularly difficult to find in the biomedical
domain. For this reason, we created a gold standard set based on
real (anonymized) user queries and the actions users subsequently
took, and named this the PubMed set.

To build the PubMed set, we collected one year's worth of search
logs and restricted the set of queries to those where users
requested the relevance order and which yielded at least 20
retrieved documents. This set contained many popular but duplicate
queries. Therefore, we merged queries and summed up user actions for
each of them. That is, for each document stored for each query, we
counted the number of times it was clicked in the retrieved set
(i.e. abstract click) and the number of times users requested
full-text articles (i.e. full-text click). We considered the queries
that appeared less than 10 times to be less informative because they
were usually very specific, and we could not collect enough user
actions for training. After this step, we further filtered out
non-informational queries (e.g. author and journal names). As the
result, 27,870 queries remained for the final set.

The last step for producing the PubMed set was to assign relevance
scores to documents for each query. We will do this based on user
clicks. It is known that click-through data is a useful proxy for
relevance judgments \cite{Joachims2002b,Agrawal2009,Xu2010}. Let
$a(D,Q)$ be the number of clicks to the abstract of a document $D$
from the results page for the query $Q$. Let $f(D,Q)$ be the number
of clicks from $D$'s abstract page to its full-text, which result
from the query $Q$. Let $\lambda \in \mathbb{R}^+$ be the boost
factor for documents without links to full-text articles. $FT (D)$
is the indicator function such that $FT (D) = 1$ if the document $D$
includes a link to full-text articles and $FT (D) = 0$ otherwise. We
can then calculate the relevance, $y$, of a document for a given
query:
\begin{eqnarray}
y(Q,D) = \mu \cdot a(Q,D) + (1-\mu) \cdot f(Q,D) + \nonumber \\
\frac{a(Q,D)}{\lambda} \cdot (1 - FT (D)),
\end{eqnarray}
\noindent $\mu$ is the trade-off between the importance of abstract
clicks and full-text clicks. The last term of the relevance function
gives a slight boost to documents without full-text links, so that
they get a better relevance (thus rank) than those for which
full-text is available but never clicked, assuming they all have the
same amount of abstract clicks. We manually tuned the parameters
based on user behavior and log analyses, and used the settings, $\mu
= 0.33$ and $\lambda = 15$.

Compared to the TREC Genomics set, the full PubMed set is much
larger, including all 27 million documents in PubMed. While the TREC
and PubMed sets share essentially the same type of documents, the
tested queries are quite different. The queries in TREC are a
question type, e.g. ``what is the role of MMS2 in cancer?" However,
the PubMed set uses actual queries from PubMed users.

In our experiments, the TREC set was used for evaluating BM25 and
the semantic measure separately and the PubMed set was used for
evaluating the learning to rank approach. We did not use the TREC
set for learning to rank due to the small number of queries. Only 62
queries and 162,259 documents are available in TREC, whereas the
PubMed set consists of many more queries and documents.

\subsection{Word Embeddings and Other Experimental Setup}

We used the skip-gram model of \textit{word2vec} \cite{Mikolov2013}
to obtain word embeddings. The alternative models such as GloVe
\cite{Pennington2014} and FastText \cite{Bojanowski2017} are
available, but their performance varies depending on tasks and is
comparable to \textit{word2vec} overall \cite{Muneeb2015,Cao2017}.
\textit{word2vec} was trained on titles and abstracts from over 25
million PubMed documents. Word vector size and window size were set
to 100 and 10, respectively. These parameters were optimized to
produce high recall for synonyms \cite{Yeganova2016}. Note that an
independent set (i.e. synonyms) was used for tuning
\textit{word2vec} parameters, and the trained model is available
online (\url{https://www.ncbi.nlm.nih.gov/IRET/DATASET}).

For experiments, we removed stopwords from queries and documents.
BM25 was chosen for performance comparison and the parameters were
set to $k=1.9$ and $b=1.0$ \cite{Lin2007}. Among document ranking
functions, BM25 shows a competitive performance \cite{Trotman2014}.
It also outperforms co-occurrence based word embedding models
\cite{Atreya2011,Nalisnick2016}. For learning to rank approaches,
70\% of the PubMed set was used for training and the rest for
testing. The RankLib library
(\url{https://sourceforge.net/p/lemur/wiki/RankLib}) was used for
implementing LambdaMART and the PubMed experiments.

\subsection{TREC Experiments}

Table \ref{tab:02} presents the average precision of \textit{tf-idf}
(TFIDF), BM25, word vector centroid (CENTROID) and our embedding
approach on the TREC dataset. Average precision \cite{Turpin2006} is
the average of the precisions at the ranks where relevant documents
appear. Relevance judgements in TREC are based on the pooling method
\cite{Manning2008}, i.e. relevance is manually assessed for top
ranking documents returned by participating systems. Therefore, we
only used the documents that annotators reviewed for our evaluation
\cite{Lu2009}.

\begin{table}
\centering
\begin{tabular}{ccc}
\hline \bf Method & \bf TREC 2006 & \bf TREC 2007 \\ \hline
TFIDF & 0.3018 & 0.2375 \\
BM25 & 0.3136 & 0.2463 \\
CENTROID & 0.2363 & 0.2459 \\
SEM & 0.3732 & 0.2601 \\
\hline
\end{tabular}
\caption{Mean average precision of \textit{tf-idf} (TFIDF), BM25,
word vector centroid (CENTROID) and our semantic approach (SEM) on the TREC set.\label{tab:02}}
\end{table}

As shown in Table \ref{tab:02}, BM25 performs better than TFIDF and
CENTROID. CENTROID maps each query and document to a vector by
taking a centroid of word embedding vectors, and the cosine
similarity between two vectors is used for scoring and ranking
documents. As mentioned earlier, this approach is not effective when
multiple topics exist in a document. From the table, the embedding
approach boosts the average precision of BM25 by 19\% and 6\% on
TREC 2006 and 2007, respectively. However, CENTROID provides scores
lower than BM25 and SEM approaches.

Although our approach outperforms BM25 on TREC, we do not claim that
BM25 and other traditional approaches can be completely replaced
with the semantic method. We see the semantic approach as a means to
narrow the gap between words in documents and those in queries (or
users' intentions). This leads to the next experiment using our
semantic measure as a feature for ranking in learning to rank.

\subsection{PubMed Experiments}

For the PubMed dataset, we used learning to rank to combine BM25 and
our semantic measure. An advantage of using learning to rank is its
flexibility to add more features and optimize performance by
learning their importance. PubMed documents are semi-structured,
consisting of title, abstract and many more fields. Since our
interest lies in text, we only used titles and abstracts, and
applied learning to rank in two different ways: 1) to find
semantically closest words in titles (BM25 +
SEM\textsubscript{Title}) and 2) to find semantically closest words
in abstracts (BM25 + SEM\textsubscript{Abstract}). Although our
semantic measure alone produces better ranking scores on the TREC
set, this does not apply to user queries in PubMed. It is because
user queries are often short, including around three words on
average, and the semantic measure cannot differentiate documents
when they include all query words.

\begin{table*}
\centering
\begin{tabular}{cccc}
\hline \bf Method & \bf NDCG@5 & \bf NDCG@10 & \bf NDCG@20 \\ \hline
BM25 & 0.0854 & 0.1145 & 0.1495 \\
BM25 + SEM\textsubscript{Title} & 0.1048 (22.72\%) &  0.1427 (24.59\%) & 0.1839 (23.03\%) \\
BM25 + SEM\textsubscript{Abstract} & 0.0917 (7.38\%) & 0.1232 (7.57\%) & 0.1592 (6.51\%) \\
\hline
\end{tabular}
\caption{NDCG scores for BM25 and learning to rank (BM25 + SEM) search results.
We used two fields from PubMed documents for the learning to rank approach.
``Title" and ``Abstract" mean only words from titles and abstracts were used to compute semantic scores, respectively.
The scores in parentheses show the improved ratios of BM25 + SEM to BM25 ranking.
\label{tab:03}}
\end{table*}

Table \ref{tab:03} shows normalized discounted cumulative gain
(NDCG) scores for top 5, 10 and 20 ranked documents for each
approach. NDCG \cite{Burges2005} is a measure for ranking quality
and it penalizes relevant documents appearing in lower ranks by
adding a rank-based discount factor. In the table, reranking
documents by learning to rank performs better than BM25 overall,
however the larger gain is obtained from using titles (BM25 +
SEM\textsubscript{Title}) by increasing NDCG@20 by 23\%. NDCG@5 and
NDCG@10 also perform better than BM25 by 23\% and 25\%,
respectively. It is not surprising that SEM\textsubscript{Title}
produces better performance than SEM\textsubscript{Abstract}. The
current PubMed search interface does not allow users to see
abstracts on the results page, hence users click documents mostly
based on titles. Nevertheless, it is clear that the abstract-based
semantic distance helps achieve better performance.

After our experiments for Table \ref{tab:03}, we also assessed the
efficiency of learning to rank (BM25 + SEM\textsubscript{Title}) by
measuring query processing speed in PubMed relevance search. Using
100 computing threads, 900 queries are processed per second, and for
each query, the average processing time is 100 milliseconds, which
is fast enough to be used in the production system.

\section{Conclusion}

We presented a word embedding approach for measuring similarity
between a query and a document. Starting from the Word Mover's
Distance, we reinterpreted the model for a query-document search
problem. Even with the $Q \rightarrow D$ flow only, the word
embedding approach is already efficient and effective. In this
setup, the proposed approach cannot distinguish documents when they
include all query words, but surprisingly, the word embedding
approach shows remarkable performance on the TREC Genomics datasets.
Moreover, applied to PubMed user queries and click-through data, our
semantic measure allows to further improves BM25 ranking
performance. This demonstrates that the semantic measure is an
important feature for IR and is closely related to user clicks.

While many deep learning solutions have been proposed recently,
their slow training and lack of flexibility to adopt various
features limit real-world use. However, our approach is more
straightforward and can be easily added as a feature in the current
PubMed relevance search framework. Proven by our PubMed search
results, our semantic measure improves ranking performance without
adding much overhead to the system.

\section*{Acknowledgments}

This research was supported by the Intramural Research Program of
the NIH, National Library of Medicine.

\bibliography{arXiv_2017}
\bibliographystyle{emnlp2016}

\end{document}